\documentclass[runningheads]{llncs}
\pdfoutput=1
\usepackage[backref=page]{hyperref}
\usepackage{amsmath}
\usepackage{url}
\usepackage{bm}
\usepackage{graphicx}
\usepackage{subfigure}
\usepackage{comment}
\usepackage{amsmath,amssymb} % define this before the line numbering.
\usepackage{color}
\usepackage{multirow}
\usepackage{booktabs}
\usepackage{array}
\usepackage{lineno,hyperref}
\usepackage[misc]{ifsym}
\usepackage[linesnumbered,boxed,ruled,commentsnumbered]{algorithm2e}%%算法包，注意设置所需可选项

\begin{document}
% \linenumbers
\pagestyle{headings}
\mainmatter
\def\ECCVSubNumber{6114}  % Insert your submission number here

%\title{Projection Perturbation to Improve the Transferability of Adversarial Examples} % Replace with your title
\title{Patch-wise Attack for Fooling Deep Neural Network}

\begin{comment}
\titlerunning{ECCV-20 submission ID \ECCVSubNumber} 
\authorrunning{ECCV-20 submission ID \ECCVSubNumber} 
\author{Anonymous ECCV submission}
\institute{Paper ID \ECCVSubNumber}
\end{comment}
%******************

%\begin{comment}
\titlerunning{Patch-wise Attack}

\author{Lianli Gao\inst{1} \and
Qilong Zhang\inst{1} \and
Jingkuan Song\inst{1} \and
Xianglong Liu\inst{2} \and
Heng Tao Shen\inst{1}\thanks{Corresponding author}}
\index{Shen, Heng Tao}
\authorrunning{L.Gao, Q.Zhang et al.}
\institute{ Center for Future Media and School of Computer Science and Engineering, University of Electronic Science and Technology of China, China 
	\and Beihang University, China
\\
\email{qilong.zhang@std.uestc.edu.cn}}
%\end{comment}
%******************
\maketitle

\begin{abstract}
By adding human-imperceptible noise to clean images, the resultant adversarial examples can fool other unknown models.
Features of a pixel extracted by deep neural networks (DNNs) are influenced by its surrounding regions, and different DNNs generally focus on different discriminative regions in recognition. Motivated by this, we propose a patch-wise iterative algorithm -- a black-box attack towards mainstream normally trained and defense models, which differs from the existing attack methods manipulating pixel-wise noise.
In this way, without sacrificing the performance of white-box attack, our adversarial examples can have strong transferability. Specifically, we introduce an amplification factor to the step size in each iteration, and one pixel's overall gradient overflowing the $\epsilon$-constraint is properly assigned to its surrounding regions by a project kernel. 
Our method can be generally integrated to any gradient-based attack methods.
Compared with the current state-of-the-art attacks, we significantly improve the success rate by 9.2\% for defense models and 3.7\% for normally trained models on average. Our code is available at \url{https://github.com/qilong-zhang/Patch-wise-iterative-attack}

\end{abstract}

\section{Introduction}
\label{Introduce}
%我这个换行的行间距比较小怎么办
%physical world~\cite{ref_article3,ref_article4}, 第三篇不确定算不算	
In recent years, Deep neural networks (DNNs)~\cite{ref_article29,ref_article30,ref_article28,ref_article27,li2019beyond,li2019learnable} have made great achievements. However, the adversarial examples~\cite{ref_article2} which are added with human-imperceptible noise can easily fool the state-of-the-art DNNs to give unreasonable predictions. This raises security concerns about those machine learning algorithms. In order to understand DNNs better and improve its robustness to avoid future risks~\cite{ref_article4}, it is necessary to investigate the defense models, and meanwhile the generation of adversarial examples, e.g., \cite{xu2019}.
% transferability那边需要再完善一下使得行文流畅

Various attack methods have been proposed in these years. One of the most popular branches is gradient-based algorithms. For this branch, existing methods can be generally classified as single-step attacks and iterative attacks. In general, iterative attacks perform better than single-step attacks in the white-box setting. But in the real world, attackers usually cannot get any information about the target model, which is called the black-box setting. In this case, single-step attacks always have a higher transferability than iterative attacks at the cost of poor performance of substitute models. 
%E.g., adversarial examples which are generated by the translation-invariant attack method~\cite{ref_article12} are less sensitive to the discriminative regions \cite{ref_article34} of the white-box models, but it demonstrates a high success rate of attack towards black-box defense models. 
To sum up, the essential difference between the two sub-branches is the number of iterations. Iterative attacks require multiple iterations to obtain the final perturbation noise, and hence there is a risk of getting stuck in the local optimum during the iterations, reducing the transferability. While single-step attack methods only update once, which is easy to underfit but really improve the generalizability. 

Moreover, with the development of attack methods, several adversarial examples have been applied to the physical world~\cite{Liu2020Biasbased,ref_article4,ref_article19,ref_article17,ref_article18}. This has raised public concerns about AI security. Consequently, a lot of defense methods are proposed to tackle this problem. Lin \textit{et at.} \cite{lin2019} propose defensive quantization method to defend adversarial attacks while maintain the efficiency. Guo \textit{et al.}~\cite{ref_article21} use bit-depth reduction, JPEG compression~\cite{ref_article22}, total variance minimization~\cite{ref_article23}, and image quilting~\cite{ref_article24} to preprocess inputs before they are feed to DNNs. Tramèr \textit{et al.}~\cite{ref_article25} use {ensemble adversarial training} to improve the robustness of models. Furthermore, Xie \textit{et al.}~\cite{ref_article26} add feature denoising module into adversarial training, and the resultant defense models demonstrate greater robustness in both white-box and black-box attack settings. 
\begin{figure}
	\centering
	\includegraphics[height=5cm]{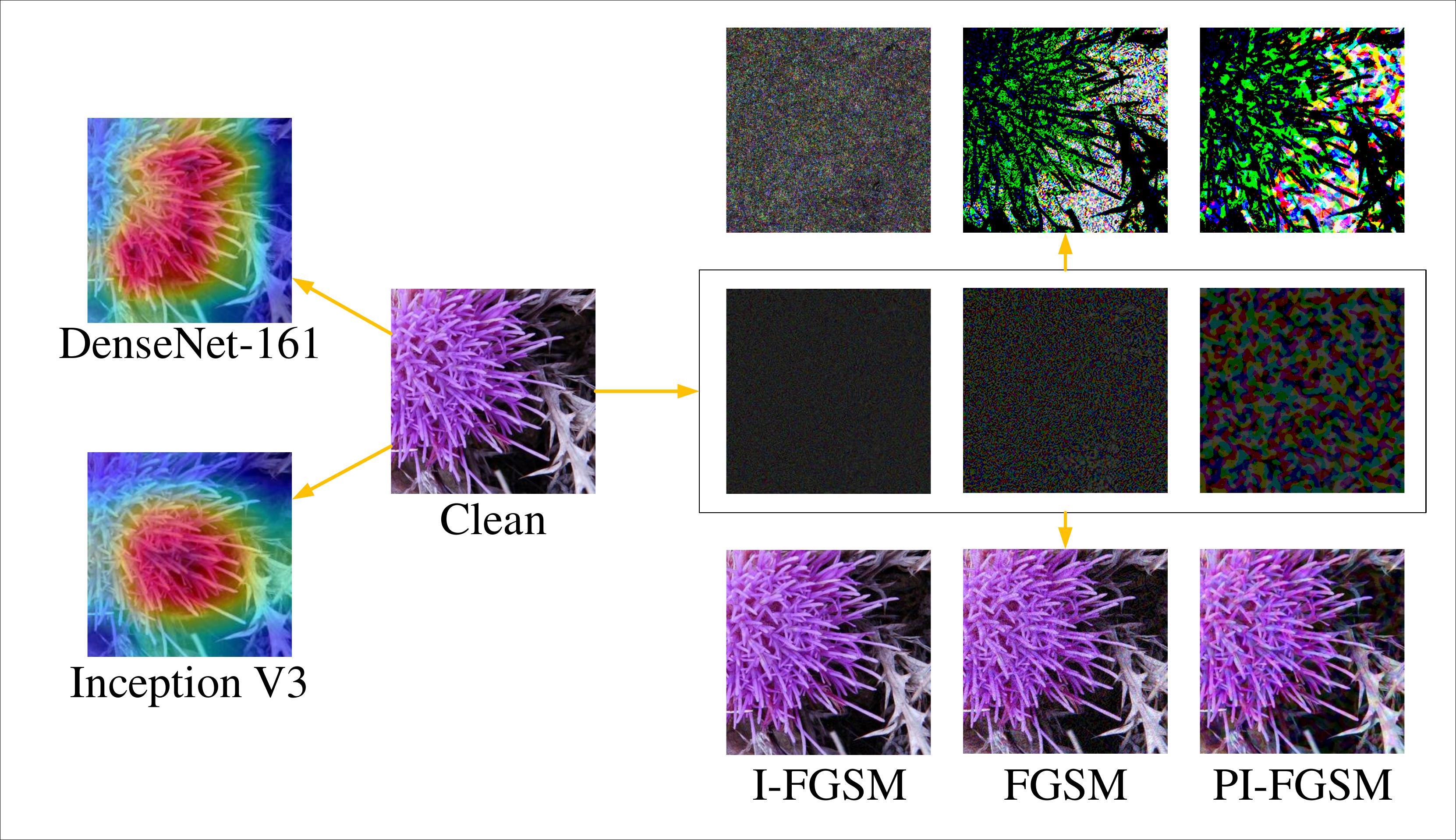}
	\caption{We show the adversarial examples generated by FGSM~\cite{ref_article8}, I-FGSM \cite{ref_article9} and our method PI-FGSM for Inception V3 model \cite{ref_article27} respectively. The maximum perturbation $\epsilon$ is limited to 16. On the left side of the figure, we use Inception V3 \cite{ref_article27} and DenseNet-161 \cite{ref_article30} to show the Gradient-weighted Class Activation Mapping (Grad-CAM) \cite{ref_article33} of the ground-truth label, and on the right side of the figure, we divide it into three parts. \textbf{Top row}: the adversarial noise patch map (we define it in Sec. \ref{map}). \textbf{Middle row}: the adversarial noise. \textbf{Bottom row}: the adversarial examples. Our PI-FGSM can generate adversarial noise which has the same clustering property as the activation map and also well covers the different discriminative regions.}
	\label{fig1}
\end{figure}
%Despite of the great progress for adversarial attack, it remains challenging, especially with the proposal of several defense %models~\cite{ref_article21,ref_article25,ref_article26}. 
To generate more effective adversarial examples, it is necessary to study the properties of intrinsic classification logic of the DNNs. Here we use {class activation mapping} \cite{ref_article34} to analyze it. Zhou \textit{et al.} \cite{ref_article34} observed that the discriminative regions always vary across predicted labels. Recent research \cite{ref_article12} also showed that different models focus on different discriminative regions in recognition, and the defense models generally focus on larger discriminative regions than the normally trained models. 
The discriminative regions are often clustered together, as shown in Fig.\ref{fig1}.
Therefore, only adding pixel-wise noise may hinder the transferability of adversarial examples across different DNNs.
Based on these observations, we argue that in addition to reducing the responsiveness of the ground-truth regions while activating the regions of any other categories, crafting perturbation with the characteristic of aggregation in the above-discussed regions is also important.
To that end, we study the advantages and disadvantages of single-step and iterative attacks \cite{ref_article12,ref_article8,ref_article9}, and we argue that linear nature of DNNs \cite{ref_article8} does exist to some extent. Therefore we amplify step size with a fixed factor to analyze the effect of step size on transferability. Besides, we rethink the weakness of direct clipping operation which discards partial gradient information. To alleviate this problem, we reuse the cut noise and apply a heuristic project strategy to reduce the side effects of direct clipping as well as generating patch-wise noise. Finally, combined with the fast sign gradient method, we propose the \textbf{Patch-wise Iterative Fast Gradient Sign Method (PI-FGSM)} to generate strongly transferable adversarial examples. 
The visualization results in Fig. \ref{fig1} also demonstrate our approach. Compared with other methods, the noise generated by our PI-FGSM has obvious aggregation characteristics and can better cover varied discriminative regions of different DNNs.
Our major contributions can be summarized as:
1) We propose a novel patch-wise attack idea named PI-FGSM. Compared with existing methods manipulating pixel-wise noise, our approach can have the advantage of both single-step and iterative attacks, i.e., improving the transferability without sacrificing the performance of the substitute model.
2) Technically, based on the mature gradient-based attack pipeline, we adopt an amplification factor and project kernel to generate more transferable adversarial examples by patch-wise noise. Our method can be generally integrated to any iteration-based attack methods;
and 3) Extensive experiments on ImageNet show that our method significantly outperforms the state-of-the-art methods, and improves the success rate by 9.2\% for defense models and 3.7\% for normally trained models on average in the black-box setting. 

\section{Related work}
In this section, we briefly analyze the exiting adversarial attack methods, from the perspectives of classification of adversarial examples, attack setting, and ensemble strategy.
\subsection{Adversarial Examples}
%讲讲生成对抗样本的条件 扰动小 啊这类
Adversarial examples are first discovered by Szegedy \textit{et al.}\cite{ref_article2}, which only added subtle perturbation to the original image but can mislead the DNNs to make an unreasonable prediction with unbelievably high confidence. To make matters worse, adversarial examples also exist in physical world \cite{ref_article4,ref_article19,ref_article3}, which raises security concerns about DNNs. Due to the vulnerability of DNNs, a large number of attack methods have been proposed and applied to various fields of deep learning in recent years, e.g., object detection and semantic segmentation \cite{DBLP:conf/iccv/XieWZZXY17}, embodied agents \cite{Liu2020Spatiotemporal}, and speech recognition \cite{DBLP:journals/corr/CisseANK17}. To make our paper more focused, we only analyze adversarial examples in the image classification task.

\subsection{Attack Settings}
%In this setting, the adversary even does not need to observe the output of the target model on any chosen inputs but still gives a powerful attack.
In this section, we describe three common attack settings. The first is the white-box setting where the adversary can get the full knowledge of the targeted models, thus obtaining accurate gradient information to update adversarial examples. The second is the semi-black-box setting where the output of the targeted model is available but model parameters are still unknown. For example, Papernot \textit{et al.}~\cite{ref_article5} train a local model with many queries to substitute for the target model. Ilyas \textit{et al.}~\cite{ref_article6} propose the variant of NES~\cite{ref_article7} to generate adversarial examples with limited queries. The rest is the black-box setting, where the adversary generally cannot access the target model and adversarial examples are usually generated for substitute models without exception. That is why transferability plays a key role in this setting. Recently, the black-box attack is a hot topic and many excellent works have been proposed. Xie \textit{et al.}~\cite{ref_article11} apply random transformations to the input images at each iteration to improve transferability. Dong \textit{et al.}~\cite{ref_article10} propose a momentum-based iterative algorithm to boost adversarial attacks. Besides, the adversarial examples which are crafted by their translation-invariant attack method~\cite{ref_article12} can evade the defenses with effect. However, the above works cannot generate powerful patch-wise noise because they generally take valid gradient information into account. In this paper, our goal is crafting efficient patch-wise noise to improve the transferability of adversarial examples in the black-box setting.
\subsection{Ensemble Strategy}
There are two well-known but totally different strategies for this topic.
One strategy uses an ensemble of legitimate examples to update only one universal adversarial perturbation. Moosavi-Dezfooli \textit{et al.}~\cite{ref_article15} propose an iterative attack method to generate such perturbations which cause almost all images sampled from the data distribution to be misclassified. Another strategy uses an ensemble of models to get a better estimate of the gradient information. Liu \textit{et al.}~\cite{ref_article16} propose novel ensemble-based approaches that attacking multiple models to generate adversarial examples. In this way, the adversarial examples are less likely to get stuck in the local optimum of any specific model, thus improving transferability.

%\subsection{Defend Methods}
%真实世界、路标、自动驾驶
%With the development of attack methods, several adversarial examples have been applied to the physical world~\cite{ref_article17,ref_article18,ref_article19,ref_article20}. This has raised public concerns about AI security. Consequently, a lot of defense methods are proposed to solve this problem. Guo \textit{et al.}~\cite{ref_article21} use bit-depth reduction, JPEG compression~\cite{ref_article22}, total variance minimization~\cite{ref_article23}, and image quilting~\cite{ref_article24} to preprocess inputs before they feed to DNN. Tramèr \textit{et al.}~\cite{ref_article25} use \textit{ensemble adversarial training} to improve the robustness of models. Furthermore, Xie \textit{et al.}~\cite{ref_article26} add feature denoising module into adversarial training, and the feature denoising networks have a very strong adversarial robustness in both white-box and black-box attack settings. 

\section{Methodology}
In this section, we describe our algorithm in detail. Let $x^{clean}$ denote a clean example without any perturbation and $y$ denote the corresponding ground-truth label. We use $f(x)$ to denote the prediction label of DNNs, and $x^{noise}$ to denote the human-imperceptible perturbation. The adversarial example $x^{adv} = x^{clean} + x^{noise}$ is visually indistinguishable from $x^{clean}$ but misleads the classifier to give a high confidence of a wrong label. 
In this paper, we focus on untargeted black-box attack, i.e., $f(x^{adv})\neq y$. And the targeted version can be simply derived. To measure the perceptibility of adversarial perturbations, we follow previous works \cite{ref_article10,ref_article12,ref_article11}  and use $l_{\infty}$-norm here. Namely, we set the max adversarial perturbation $\epsilon$, i.e., we should keep $||x^{clean}-x^{adv}||_{\infty} \leq \epsilon$. To generate our adversarial examples, we should maximize the cross-entropy loss $J(x^{adv}, y)$ of the substitute models. Our goal is to solve the following constrained optimization problem:
\begin{equation}
\underset{x^{adv}}{\arg \max} J(x^{adv}, y), \qquad \qquad  s.t.\ ||x^{clean}-x^{adv}||_\infty \leq \epsilon.
\end{equation}
Due to the black-box setting, the adversary does not allow to analytically compute the target models' gradient $\nabla J(x, y)$. In the majority of cases, they use the information of substitute models (i.e., official pre-trained models) to generate adversarial examples. Therefore, it is very important to improve the transferability of adversarial examples so that they still fool the black-box models successfully.

\subsection{{Development of Gradient-based Attack Methods }}
In this section, we give a brief introduction of some excellent works which are based on the {gradient sign method}.
\begin{itemize}
	\item \textit{Fast Gradient Sign Method (FGSM)}: Goodfellow \textit{et al.}~\cite{ref_article8} argue that the vulnerability of DNN is their linear nature. Consequently they update the adversarial example by: 
	\begin{equation}
	x^{adv} = x^{clean} + \epsilon \cdot sign(\nabla_x J(x^{clean}, y)).
	\end{equation}
	where $sign(\cdot)$ indicates the sign operation.
	%This method is the first member of this family, it sacrifice the performance in white-box scenario for higher transferability in black-box scenario. 
	
	\item \textit{Iterative Fast Gradient Sign Method (I-FGSM)}: Kurakin \textit{et al.}~\cite{ref_article9} use a small step size $\alpha$ to iteratively apply the gradient sign method multiple times. This method can be written as:
	\begin{equation} 
	x^{adv}_{t+1} = Clip_{x^{clean}, \epsilon}\{x^{adv}_{t} + \alpha \cdot sign(\nabla_x J(x^{adv}_t, y))\}.
	\label{I-FGSM}
	\end{equation} 
	where $Clip_{x^{clean}, \epsilon}$ denotes element-wise clipping, aiming to restrict $x^{adv}$ within the $l_\infty$-bound of $x^{clean}$. 
	
	\item \textit{Momentum Iterative Fast Gradient Sign Method (MI-FGSM)}: Dong \textit{et al.}~\cite{ref_article10} use momentum term to stabilize update directions. It can be expressed as:
	\begin{equation}
	g_{t+1} = \mu \cdot g_t + \frac{\nabla_x J(x^{adv}_t, y)}{||\nabla_x J(x^{adv}_t, y) ||_1},~~
	x^{adv}_{t+1} = x^{adv}_t + \alpha \cdot sign(g_{t+1}).
	\label{eq.mifgsm}
	\end{equation}
	where $g_t$ is cumulative gradient, and $\mu$ is the decay factor.
	
	\item \textit{Diverse Input Iterative Fast Gradient Sign Method (DI$^2$-FGSM)}: Xie \textit{et al.}~\cite{ref_article11} apply diverse input patterns to improve the transferability of adversarial examples. With the replacement of Eq. (\ref{I-FGSM}) by:
	\begin{equation}
	x^{adv}_{t+1} = Clip_{x^{clean}, \epsilon}\{x^{adv}_{t} + \alpha \cdot sign(\nabla_x J(D(x^{adv}_t), y))\}.
	\end{equation}
	where $D(x)$ is random transformations to the input $x$. For simplicity, we use DI-FGSM later.
	
	\item \textit{Translation-Invariant Attack Method}: Dong \textit{et al.}~\cite{ref_article12} convolve the gradient with the pre-defined kernel $W$ to generate adversarial examples which are less sensitive to the discriminative regions of the substitute model. For {TI-FGSM}, it is only updated in one step:
	\begin{equation}
	x^{adv} = x^{clean} + \epsilon \cdot sign(W *\nabla_x J( x^{adv}_t, y)).
	\end{equation}
	and TI-BIM is its iterative version.
\end{itemize}

\subsection{Patch-wise Iterative Fast Gradient Sign Method}
%对比DI TI MI I FGSM这些，写在介绍那边吧 

%Although many excellent works have been proposed, we still think that the non-targeted attacks in the black-box setting are not strong enough. 
In this section, we elaborate our method in details. We first introduce our motivations in {Sec. \ref{map} and Sec. \ref{box}}. In Sec. \ref{sol}, we will describe our solution. 
%In Sec. \ref{ext}, we will simply introduce the $L_2$-norm attack extension of our projection strategy.

\subsubsection{Patch Map.}
\label{map}
Natural images are generally made up of smooth patches \cite{DBLP:conf/cvpr/MahendranV15} and the discriminative regions are usually focused on several patches of them.
However, as demonstrated in Fig. \ref{fig1}, different DNNs generally focus on different discriminative regions, but these regions usually contain clustered pixels instead of scattered ones. Besides, Li et al. \cite{region} have demonstrated that regionally homogeneous perturbations are strong in attacking defense models, which is especially helpful to learn transferable adversarial examples in the black-box setting. 
For this reason, we believe that noises with the characteristic of aggregation in these regions are more likely to attack successfully because they perturb more significant information. 
To better view the adversarial noise $x^{noise}$, we take the absolute value of $x^{noise}$ to define its patch map $x^{map}$\footnote{Pixel values of a valid image are in [0, 255]. If the values are more than 255, they will be modified into 0 {for} ``uint8'' type, to give better contrast.}, which is done by:
%\footnote{https://pypi.org/project/PIL/}
\begin{equation}
\label{F}
x^{map}=|x^{noise}| \times \frac{256}{\epsilon}.
\end{equation}
%the black regions denote the pixel absolute values of three channels(RGB) are zero or all reach $\epsilon$, and other colors are not.
As shown in Fig. \ref{fig1}, compared with the patch map of I-FGSM, and FGSM, our PI-FGSM can generate the noise with more obvious aggregation characteristics. %visualization of FGSM can clearly see the connection between the noise and the original image. This phenomenon can also proves that the noise with aggregation characteristics will have better transferability. But if zooming in on the patch map, we can find that $x^{map}$ generated by FGSM is still somewhat sparse, e.g., the green regions. 

\subsubsection{Box Constraint.}
\label{box}
To the best of our knowledge, almost all iterative gradient-based methods apply \textit{projected gradient descent} to ensure the perturbation within the box. Although this method can improve the generalization of adversarial examples to some extent~\cite{ref_article15}, it also has certain limitations. Let us take the dot product $D(\cdot)$ as an example:
\begin{equation}
D(x_t^{adv}) = wx_t^{adv}+b, \qquad \qquad D^{'}(x_t^{adv}) = w.
\end{equation}
where $w$ denotes a weight vector and $b$ denotes the bias. Then we add a noise $\alpha w$ to update $x_t^{adv}$:
%\begin{equation}
%D(x_t+\alpha \cdot w)=D(x_t)+\alpha w^2 
%\end{equation}
\begin{equation}
\label{14}
D(Clip_{x^{clean}, \epsilon}\{x_t^{adv}+\alpha w\}) \approx D(x_t^{adv}) + \alpha_2 w^2.
\end{equation}
If $x_{t}^{adv} +\alpha w$ excess the $\epsilon$-ball of original image $x^{clean}$, the result is Eq.(\ref{14}). Obviously, $\alpha_2 \textless \alpha$ due to element-wise clipping operation. If we adopt this strategy directly, we will waste some of the gradient information and change the input unexpectedly. 
%Rosen投影梯度法能加进去吗
%To perform a more transferable attack, we propose a novel heuristic projection strategy. Our inspiration comes from \textit{Rosen Project Gradient Method}\cite{ref_article35}: {\color{red} By projecting the gradient direction when the iteration point is on the edge of the feasible region}, the method ensures the iteration point remains within the feasible region after updating. However, performing this method is a little complex and needs additional computational cost. So we want to take some heuristic strategies to apply this idea. 
%our heuristic project strategy guides excess noise to their feasible descent direction:

\subsubsection{Our Method.}
\label{sol}
From the above analysis, we observe that adding noise in a patch-wise style will have better transferability than the pixel-wise style. Also, the element-wise clipping operation of existing gradient-based attack methods will lose part of the gradient information and lead to unexpected changes.   
Therefore, we propose our method, which follows the mature gradient-based attack pipeline and tackles the above issues simultaneously. 

To the best of our knowledge, many recent iterative attack methods~\cite{ref_article10,ref_article12,ref_article11} set step size $\alpha = \epsilon/T$, where $T$ is the total number of iterations. In such a setting, we do not need the element-wise clipping operation, and the adversarial examples can finally reach the $\epsilon$ bound of $x^{clean}$. This seems like a good way to get around the above problem of direct clipping, but we notice that single-step attacks often outperform iterative attacks in the black-box setting. To study the transferability with respect to the step size setting, we make a tradeoff between single large step and iterative small step by setting it to $\epsilon/T\times \beta$, where $\beta$ is an amplification factor.
%In this paper, we set step size to $\epsilon/T\times \beta$.

The results in Fig. \ref{fig3} show that iterative approaches with a large amplification factor will help to avoid getting stuck in poor local optimum, thus demonstrating a stronger attack towards black-box models. One possible reason is that attacks with an amplification factor increase each element's value of the resultant perturbation, thus providing a higher probability of misclassification due to the linear assumption of Goodfellow \textit{et al.} \cite{ref_article8}. However, simply increasing the step size does not get around the disadvantages of direct clipping operation, because the excess noise would be eliminated. 

Therefore, we propose a novel heuristic project strategy to solve this problem. Our inspiration comes from {Rosen Project Gradient Method} \cite{ref_article35}: by projecting the gradient direction when the iteration point is on the edge of the feasible region, the method ensures the iteration point remains within the feasible region after updating. However, performing this method is a little complex and needs additional computational cost. Hence we take a heuristic strategy to apply this idea: just projecting the excess noise into the surrounding field. We argue that the part of the noise vector which is more easy to break $\epsilon$-ball limitation has a higher probability of being in the highlighted area of discriminative regions. Our strategy can simply reuse the noise to increase the degree of aggregation in these regions without additional huge computational cost.
%Furthermore, this setting can outperform its single-step counterpart sometimes.
% 两个决策边界
%\begin{figure}
%	\centering
%	\includegraphics[height=4cm]{images/line.png}
%	\caption{The trajectory of different step size. We use the background contour map to represent the cross-entropy loss of the substitute model roughly. The deeper the color, the lower the loss. The yellow solid circle represents a clean image, while others represent its adversarial version. In this figure, the solid blue, red, black lines represent the trajectory of FGSM, iterative methods of small step size and big step size respectively.}
%	\label{fig2}
%\end{figure}
\IncMargin{1em} % 使得行号不向外突出 
\begin{algorithm}[t]
	\DontPrintSemicolon
	\SetAlgoNoLine % 不要算法中的竖线
	\SetKwInOut{Input}{\textbf{Input}}\SetKwInOut{Output}{\textbf{Output}}
	% clipped noise， additional noise换个说法吧
	\Input{The cross-entropy loss function $J$ of our substitute models; iterations $T$; $L_{\infty}$ constraint $\epsilon$; project kernel $W_p$; amplification factor $\beta(\geq 1)$; project factor $\gamma$; a clean image $x^{clean}$ (Normalized to [-1,1]) and the corresponding groud-truth label $y$; 
	}
	\Output{The adversarial example $x^{adv}$;
	}
	Initialize cumulative amplification noise $a_0$ and cut noise $C$ to 0;\quad\\
	$x^{adv}_0 = x^{clean}$;
	
	\For{$t \leftarrow 0$ \KwTo $T$}
	{
		Calculate the gradient $\nabla_x J( x^{adv}_t, y)$;\quad\\
		
		$a_{t+1} = a_{t} + \beta \cdot \frac{\epsilon}{T} \cdot sign(\nabla_x J( x^{adv}_t, y))$; \tcp*{Update $a_{t+1}$}
		\eIf{$||a_{t+1}||_{\infty} \ge \epsilon$} 
		{	
			$C = clip(|a_{t+1}|-\epsilon, 0, \infty) \cdot sign(a_{t+1})$;\quad\\
			$a_{t+1} = a_{t+1} + \gamma \cdot sign(W_p*C)$;
		}
		{
			$C = 0$;
		}
		$x^{adv}_{t+1} = Clip_{x^{clean}, \epsilon}\{x^{adv}_{t} + \beta \cdot \frac{\epsilon}{T} \cdot sign(\nabla_x J( x^{adv}_t, y)) + \gamma \cdot sign(W_p*C)\}$; 	\quad\\
		$x^{adv}_{t+1} = clip(x^{adv}_{t+1}, -1, 1)$; \tcp*{Finally clip $x^{adv}_{t+1}$ into [-1,1]}
	}
	Return $x^{adv} = x^{adv}_T$;
	\caption{PI-FGSM\label{A1}}
\end{algorithm}
\DecMargin{1em}
%, thus demonstrating a stronger attack towards black-box models
%We do amplification because if we set $\beta=1$, the cut noise $C$ will be no effect.
%And on the other hand,  Because 

The integration of patch-wise iterative algorithm and fast gradient sign method (PI-FGSM) is summarized in Algorithm \ref{A1}. Firstly, in line 5, we need to get the cumulative amplification noise $a_t$. After amplification operation, if $L_\infty$-norm of $a_t$ exceeds the threshold $\epsilon$, we update the cut noise $C$ by:
\begin{equation}
C = clip(|a_{t+1}|-\epsilon, 0, \infty) \cdot sign(a_{t+1}).
\end{equation}
where $|\cdot|$ denotes the absolute operation. Finally, unlike other methods, we add an additional project term before restricting the $L_\infty$-norm of the perturbations. Note that we do not abandon the clipping operation. Instead, we just reuse the cut noise to alleviate the disadvantages of direct clipping, thus increasing the aggregation degree of noise patches. More specifically, we update the adversarial examples by:
\begin{equation}
x^{adv}_{t+1} = Clip_{x^{clean}, \epsilon}\{ x^{adv}_{t} + \beta \cdot \frac{\epsilon}{T}\cdot sign(\nabla_x J( x^{adv}_t, y)) + \gamma \cdot sign(W_p*C)\}.
\end{equation}
where $W_p$ is a special uniform project kernel of size $k_w\times k_w$, and $sign(W_p*C)$ is cut noise's ``feasible direction''. In this paper, we simply define $W_p$ as:
\begin{equation}
W_p[i,j] =  \begin{cases}
0 ,&  i = \lfloor k_w/2 \rfloor, j =\lfloor k_w/2 \rfloor. \\

1 / (k_w^2-1) ,& else.
\end{cases}
\end{equation}
We also test other types of kernels actually, e.g., Gaussian kernel. However, experiment results show that there are no significant difference (only $\sim$1\%). Besides, the uniform kernel does not need extra parameters. Therefore we choose it finally.

\section{Experiment}
\label{exp}
\subsection{Setup}
Following the previous works \cite{ref_article10,ref_article12}, we also do our experiments on ImageNet-compatible dataset\footnote{\url{https://github.com/tensorflow/cleverhans/tree/master/examples/nips17_adversarial_competition/dataset}}, which contains 1000 images and is used for NIPS 2017 adversarial competition. We choose eleven models to do these experiments. For normally trained models, we consider Inception V3 (Inc-v3) \cite{ref_article27}, Inception V4 (Inc-v4) \cite{ref_article28}, Inception-ResNet V2 (IncRes-v2) \cite{ref_article28} ResNet152 V2 (Res-152) \cite{ref_article29} and {DenseNet 161 (Dense-161)} \cite{ref_article30}. For defense models, we include three ensemble adversarial trained models: Inc-v3$_{ens3}$, Inc-v3$_{ens4}$ and IncRes-v2$_{ens}$ \cite{ref_article25}, and three more robust models from \cite{ref_article11}: ResNet152 Baseline (Res152$_B$), ResNet152 Denoise (Res152$_D$), ResNeXt101 DenoiseAll (ResNeXt$_{DA}$) \cite{ref_article26}. For the sake of simplicity, we use NT to denote normally trained models, EAT to denote ensemble adversarial trained models and FD to denote feature denoising defense models (include Res152$_B$). Noted that the reported success rate of FD has subtracted the ratio of clean images that are predicted incorrectly by FD. 
In addition, PI-FGSM can be easily combined with other attack methods. To make the abbreviation unambiguous, we use the first character to denote the corresponding method. For instance, DPI-FGSM means the integration of DI-FGSM with PI-FGSM. Besides, we use the character ``A" to denote any other attack approach that only applies our amplification factor $\beta$, e.g., AI-FGSM denotes the resultant method that amplifies the step size ($\epsilon / T$) of I-FGSM.

In our experiment, we set equal weight when attacking an ensemble of models. The maximum perturbation $\epsilon$ is set to 16. The iteration $T$ is set to 10 for all iterative methods. For iterative methods without our amplification factor, the step size is $\epsilon / T = 1.6$. For MI-FGSM, we set the decay factor $\mu = 1.0$; for TI-FGSM and TI-BIM, we set the kernel size $k = 15$; and for DI-FGSM, we set the transformation probability $p=0.7$.  
%\footnote{In this paper, we use bigger $\beta$ to amplification noise, so we set $T=10$. It's also ok to use smaller $\beta$ with bigger $T$. What we want to say is that there is more than one choice.}
%描述图 
%接下来做结合对比

\subsection{Amplification factor}
\begin{figure}
	\centering
	\includegraphics[width=\linewidth,height=4cm]{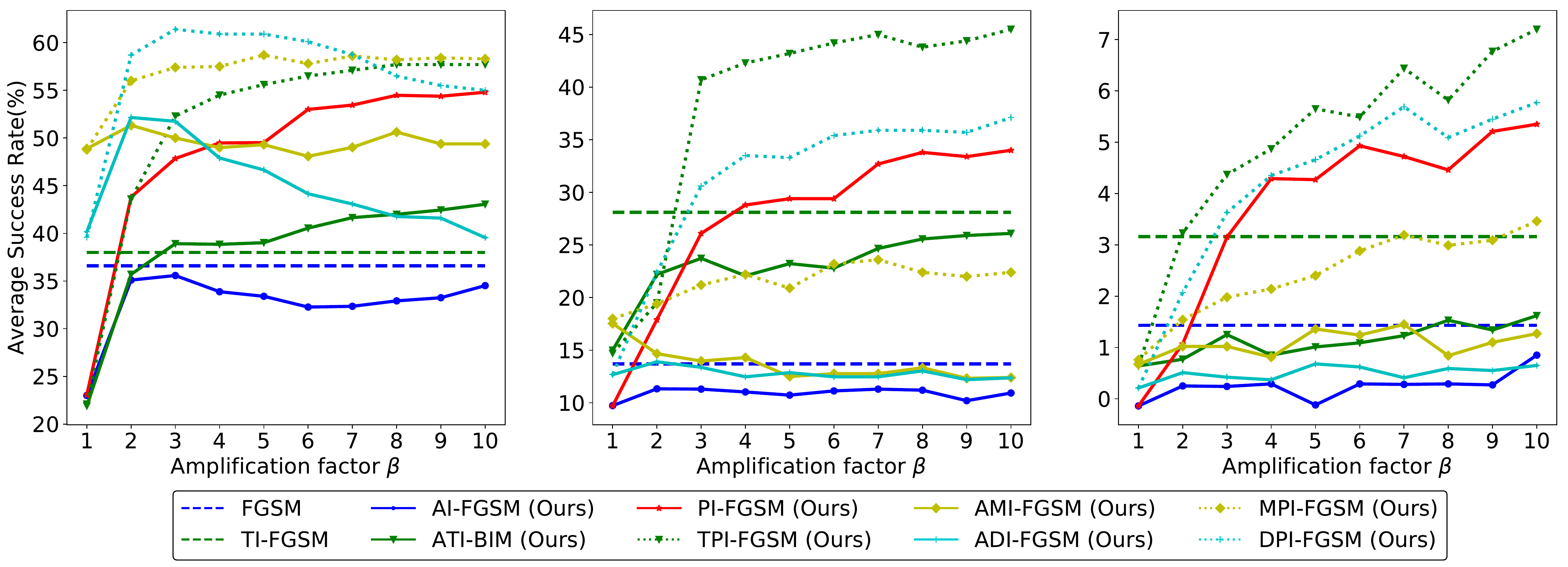}
	\caption{The average success rate(\%) of non-targeted attacks in different amplification factor $\beta$ setting. The adversarial examples are crafted for Inc-v3 by FGSM, AI-FGSM, AMI-FGSM, ATI-BIM, TI-FGSM, ADI-FGSM, PI-FGSM and their combined {versions} respectively. \textbf{Left Column}: The result of \textit{NT}, including Inc-v4, Res-152, IncRes-v2 and Dense-161 but except Inc-v3; \textbf{Middle Column}: The result of \textit{EAT}, including Inc-v3$_{ens3}$, Inc-v3$_{ens4}$ and IncRes-v2$_{ens}$; \textbf{Right Column}: The result of \textit{FD}, including {ResNeXt$_{DA}$}, Res152$_{B}$ and Res152$_{D}$.}
	\label{fig3}
\end{figure}

In this section, we calculate the success rate of different amplification factors, which are varied from 1 to 10. The results are shown in Fig.~\ref{fig3}. In general, a larger amplification factor does improve performance. Moreover, we observe that when the amplification factor is large enough, further increasing it will not bring significant improvement and may even reduce the transferability of some attacks. For example, it is better not to use large amplification factor for ADI-FGSM, because this method applies random transformations to the input images. If we set the amplification factor too large, the adversarial examples may deviate from the global optimum. In addition, a larger amplification factor has no obvious effect for AMI-FGSM, which also proves that the momentum term can stabilize the update directions and avoid getting stuck in a poor local optimum. For our method, as the amplification factor increases, the success rate will increase rapidly and outperform other approaches soon. By comparing the growth curves of AI-FGSM and PI-FGSM, we observe that our heuristic project strategy can improve the transferability by a large margin, which fully validates the effectiveness of our method.
We also examine the influence of amplification factor on the combined methods and show the results in Fig.~\ref{fig3}. It can be observed that the optimal amplification factor of combined methods is usually between the best $\beta$ of the two methods. If one of these methods' performance is negatively correlated with the amplification factor, then the combination with our method may not perform well (i.e., MPI-FGSM vs. EAT). In our experiments, the project factor $\gamma$ is set to $\epsilon/T\cdot \beta$ in most cases, therefore we only focus on the amplification factor in this ablation study, specific parameter settings will be given in the later experiments.
%Although the DPI-FGSM can perform better than PI-FGSM when attacking other models(See table. \ref{table2,table3}),
\subsection{Project Kernel Size}
%其实我想表达这些一块块的面积比较大
%For NT and EAT, we use Inc-v3, Inc-v4, Res-152, IncRes-v2 and the ensemble of these models to generate adversarial examples; For FD, we use Res152$_B$, Res152$_D$, ResNeXt$_{DA}$, an ensemble of FD and an ensemble of NT to generate adversarial examples.
\begin{figure}
	\centering
	\includegraphics[width=\linewidth,height=4.3cm]{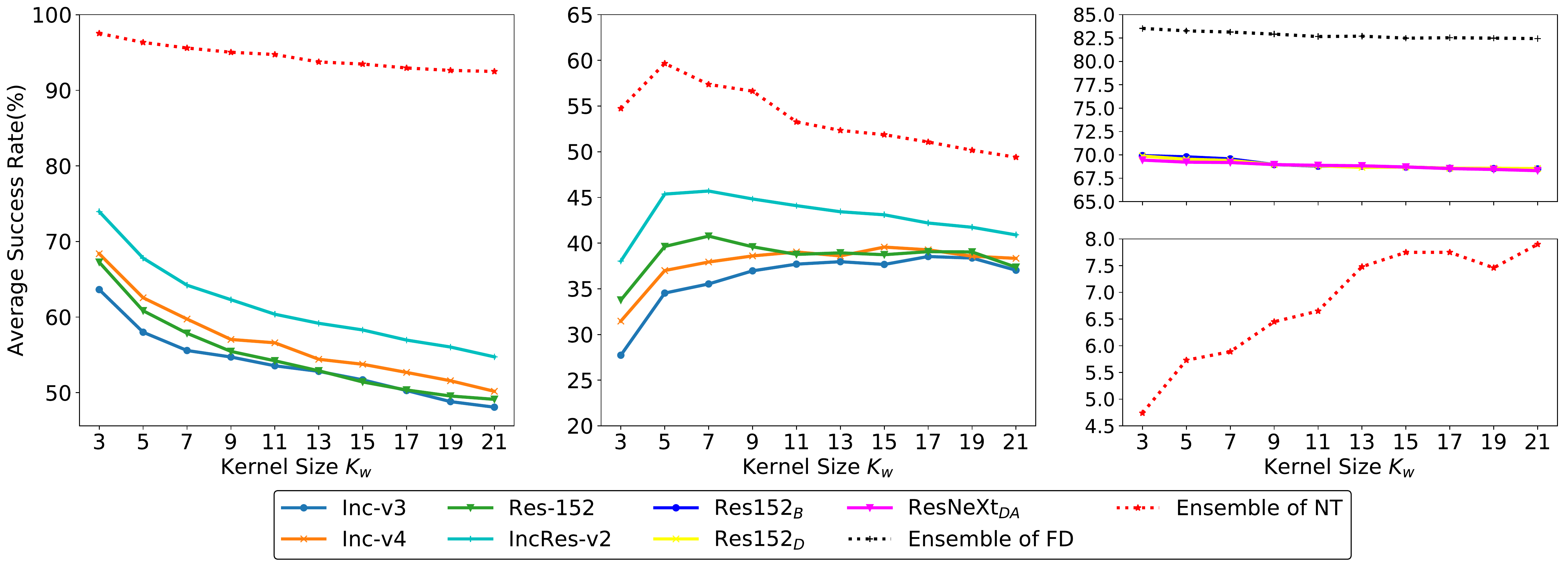}
	\caption{The average success rate(\%) of non-targeted attacks in different kernel size settings. The adversarial examples are crafted for different substitute models respectively, e.g., Inc-v3 (blue solid line). \textbf{Left Column}: The results of \textit{NT}, including Inc-v3, Inc-v4, Res-152, IncRes-v2 and Dense-161; \textbf{Middle Column}: The results of \textit{EAT}, including Inc-v3$_{ens3}$, Inc-v3$_{ens4}$ and IncRes-v2$_{ens}$; \textbf{Right Column}: The results of \textit{FD}, including ResNeXt$_{DA}$, Res152$_{B}$ and Res152$_{D}$.}
	\label{fig4}
\end{figure}
In fact, the function of the project kernel is to generate patch-wise noise. As shown in Fig. \ref{fig1}, the noise patch map of PI-FGSM is larger and regular than others. The results in Fig. \ref{fig3} also demonstrate the effectiveness of our proposed method. 
However, The size of the kernel $W_p$ play an important role for transferability. From Fig. \ref{fig4}, we find the optimal size of project kernel is different for NT, EAT and FD: 
\begin{itemize}
	\item When transferring to NT, $3\times 3$ kernel is the best. Larger kernel size always decreases the transferability in the black-box setting. 
	\item When transferring to EAT, $7\times 7$ kernel can improve the transferability obviously. But if we keep increasing the size, the success rate grows slowly and even gets worse.
	\item When transferring to FD, if adversarial examples are generated for FD, $3\times 3$ kernel is the best. And if we use NT to attack against FD, $21\times 21$ is slightly better. 
\end{itemize}
%就是想表达3x3与这个关系
Considering the difference between the NT, EAT and FD, we use $3\times 3$ kernel to attack against NT, $7\times 7$ for EAT and $21\times 21$ for FD if adversarial examples are crafted for NT. Also, we use $3\times 3$ kernel to fool FD if adversarial examples are generated for FD.
\begin{table}[h]
	\centering
	\caption{The success rate(\%) of non-targeted attacks against NT. The leftmost column models are substitute models (``*'' indicates white-box attack), the adversarial examples are crafted for them by FGSM, I-FGSM, MI-FGSM, DI-FGSM, PI-FGSM, and their combined versions respectively.}
	\resizebox{0.75\linewidth}{!}{
		\begin{tabular}{c|c|p{1.5cm}<{\centering}|p{1.5cm}<{\centering}|p{1.5cm}<{\centering}|p{1.5cm}<{\centering}|p{1.5cm}<{\centering}}
			\toprule
			\toprule
			& Attacks & Inc-v3 & Inc-v4 & Res-152 & IncRes-v2 & Dense-161 \\
			\midrule
			\midrule
			\multirow{9}[4]{*}{Inc-v3} & FGSM  & 80.9* & 38.0 & 33.1 & 33.9 & 41.4 \\
			& I-FGSM & \textbf{100.0*} & 29.6 & 19.4 & 20.3 & 20.7 \\
			& MI-FGSM & \textbf{100.0*} & 54.1 & 43.5 & 50.9 & 45.8 \\
			& DI-FGSM & 99.8* & 54.2 & 32.1 & 43.6 & 30.4 \\
			& PI-FGSM(Ours) & \textbf{100.0*} & \textbf{58.6} & \textbf{45.0} & \textbf{51.3} & \textbf{61.7} \\
			\cmidrule{2-7}          & MPI-FGSM(Ours) & \textbf{100.0*} & 63.0 & 50.6 & 60.0 & 59.0 \\
			& DPI-FGSM(Ours) & \textbf{100.0*} & 73.1 & 51.2 & 67.4 & 55.9 \\
			& DMI-FGSM & 99.9* & 78.9 & \textbf{63.9} & 75.6 & 60.7 \\
			& DMPI-FGSM(Ours) & \textbf{100.0*} & \textbf{81.8} & 63.4 & \textbf{77.1} & \textbf{63.7} \\
			\midrule
			\multirow{9}[4]{*}{Inc-v4} & FGSM  & 45.4 & 75.1* & 35.1 & 35.8 & 45.4 \\
			& I-FGSM & 43.3 & \textbf{100.0*} & 25.5 & 25.3 & 24.7 \\
			& MI-FGSM & \textbf{71.2} & \textbf{100.0*} & \textbf{52.4} & \textbf{59.0} & 51.2 \\
			& DI-FGSM & 66.6 & \textbf{100.0*} & 39.8 & 50.4 & 33.2 \\
			& PI-FGSM(Ours) & 70.7 & \textbf{100.0*} & 50.9 & 54.9 & \textbf{65.8} \\
			\cmidrule{2-7}          & MPI-FGSM(Ours) & 77.3 & \textbf{100.0*} & 56.5 & 63.7 & 64.3 \\
			& DPI-FGSM(Ours) & 84.3 & \textbf{100.0*} & 57.6 & 70.6 & 61.8 \\
			& DMI-FGSM & 89.0 & \textbf{100.0*} & \textbf{70.8} & 80.2 & 67.7 \\
			& DMPI-FGSM(Ours) & \textbf{90.4} & \textbf{100.0*} & 70.6 & \textbf{82.2} & \textbf{70.0} \\
			\midrule
			\multirow{9}[4]{*}{Res-152} & FGSM  & 41.4 & 36.4 & 82.3* & 32.0 & 45.1 \\
			& I-FGSM & 30.7 & 24.7 & 99.5* & 16.9 & 23.7 \\
			& MI-FGSM & 56.1 & 51.0 & 99.5* & 47.9 & 50.2 \\
			& DI-FGSM & 60.0 & \textbf{56.5} & 99.2* & 49.3 & 43.1 \\
			& PI-FGSM(Ours) & \textbf{63.3} & 54.4 & \textbf{99.7*} & \textbf{50.6} & \textbf{67.4} \\
			\cmidrule{2-7}          & MPI-FGSM(Ours) & 68.8 & 62.5 & \textbf{99.7*} & 59.9 & 69.0 \\
			& DPI-FGSM(Ours) & 81.0 & 77.2 & 99.6* & 75.0 & 71.6 \\
			& DMI-FGSM & 82.2 & 79.4 & 99.3* & 74.8 & 72.0 \\
			& DMPI-FGSM(Ours) & \textbf{86.1} & \textbf{83.4} & 99.5* & \textbf{82.0} & \textbf{75.2} \\
			\midrule
			\multirow{9}[4]{*}{IncRes-v2} & FGSM  & 45.9 & 39.2 & 35.7 & 68.3* & 45.6 \\
			& I-FGSM & 48.2 & 38.3 & 25.5 & \textbf{100.0*} & 27.0 \\
			& MI-FGSM & \textbf{77.6} & 68.4 & 57.0 & \textbf{100.0*} & 57.1 \\
			& DI-FGSM & 70.2 & 66.1 & 47.9 & 99.2* & 42.0 \\
			& PI-FGSM(Ours) & 76.1 & \textbf{68.8} & \textbf{58.2} & 99.9* & \textbf{69.8} \\
			\cmidrule{2-7}          & MPI-FGSM(Ours) & 80.2 & 75.5 & 63.8 & \textbf{100.0}* & 70.3 \\
			& DPI-FGSM(Ours) & 87.2 & 83.4 & 65.0 & 99.7* & 69.2 \\
			& DMI-FGSM & 86.8 & 85.6 & 76.5 & 99.1* & 70.9 \\
			& DMPI-FGSM(Ours) & \textbf{92.2} & \textbf{89.0} & \textbf{77.1} & 99.6* & \textbf{74.7} \\
			\bottomrule
	\end{tabular}}%
	\label{table1}%
\end{table}%

\subsection{Attacks vs. Normally Trained Models}
In this section, we focus on the vulnerability of NT. we set $\beta = 10$ ($\gamma=16$) for PI-FGSM and MPI-FGSM, and $\beta = 2.5, \gamma=2$ for DMPI-FGSM and DPI-FGSM. We compare PI-FGSM with FGSM, I-FGSM, MI-FGSM, DI-FGSM to verify the effectiveness of our method. Moreover, we also test the performance of the combination of different methods, e.g., MPI-FGSM. It should be noted that we do not consider TI-BIM or TI-FGSM \cite{ref_article12} here, because they focus more on attacking the defense models.

The results are shown in Tab. \ref{table1}. To sum up, compared with other attacks, our PI-FGSM can improve the success rate by \textbf{3.7\%} on average, and when we attack against Dense-161\footnote{Input size need to be [224,224,3], therefore we need resize adversarial examples whose size is [299,299,3]}, transferability can be increased by up to \textbf{17.2\%}. This is because our perturbation patches have the property of aggregation, thus reducing the impact of resizing. In particular, if we integrate PI-FGSM into other attack methods such as DMI-FGSM, we can get a much better result. For instance, the adversarial examples generated for IncRes-v2 by DMPI-FGSM can fool Inc-v3 on \textbf{92.2\%} images in the black-box setting which also demonstrates the vulnerability of NT.

\subsection{Attacks vs. Defense Models}
Our approach is especially effective for defense models. In this experiment, we use EAT \cite{ref_article25} and FD \cite{ref_article26} to examine the transferability and we do not integrate the momentum term into our proposed PI-FGSM because it may hinder the performance.

Here we study the single-model attacks firstly. In this case, we set $\beta = 10$ $(\gamma = 16)$, and the results are shown in Tab. \ref{table2}. Compared with TI-FGSM, the average success rate of our method is improved by about \textbf{9.0\%}. In particular, if we use DTPI-FGSM to attack IncRes-v2, \textbf{70.0\%} adversarial examples can fool Inc-v3$_{ens}$. Noted that the results of Tab. \ref{table2} are not our best parameter setting, because the best kernel size for a single-model attack is not the same. Here we just set the $k_w = 7$ to keep the experimental parameters consistent.

The transferability can be greatly improved when the adversarial examples are crafted for an ensemble of models at the same time \cite{ref_article16}. It is because this strategy can prevent adversarial examples from falling into a local optimum of any specific model.
In this case, we set $\beta = 5$ $(\gamma = 8)$ and the result are shown in Tab. \ref{table3}. Compared with MI-FGSM, our proposed PI-FGSM improves the performance by about \textbf{9.6\%} on average. Furthermore, compared with DTMI-FGSM \cite{ref_article12} which takes momentum into account, our DTPI-FGSM still outperform it.
\begin{table}[t]
	\centering
	\caption{The success rate(\%) of non-targeted attacks against EAT. The top row models are substitute models, we use them to generate adversarial examples by FGSM, I-FGSM, DI-FGSM, MI-FGSM, TI-FGSM, PI-FGSM and their combined versions respectively.}
	\resizebox{\textwidth}{!}{
		\begin{tabular}{c|c|c|c|c|c|c|c|c|c|c|c|c}
			\toprule
			\toprule
			& \multicolumn{3}{c|}{Inc-v3} & \multicolumn{3}{c|}{Inc-v4} & \multicolumn{3}{c|}{Res-152} & \multicolumn{3}{c}{IncRes-v2} \\
			\midrule
			\midrule
			Attacks & Inc-v3$_{ens3}$ & Inc-v3$_{ens4}$ & IncRes-v2$_{ens}$ & Inc-v3$_{ens3}$ & Inc-v3$_{ens4}$ & IncRes-v2$_{ens}$ & Inc-v3$_{ens3}$ & Inc-v3$_{ens4}$ & IncRes-v2$_{ens}$ & Inc-v3$_{ens3}$ & Inc-v3$_{ens4}$ & IncRes-v2$_{ens}$ \\
			\midrule
			\midrule
			FGSM  & 16.8 & 15.8 & 8.3 & 16.6 & 17.2 & 9.1 & 21.4 & 19.4 & 11.4 & 18.6 & 17.5 & 11.2 \\
			I-FGSM & 11.7 & 12.1 & 5.5 & 11.8 & 13.0 & 6.6 & 13.0 & 13.3 & 6.7 & 13.7 & 13.3 & 8.2 \\
			MI-FGSM & 21.9 & 21.1 & 10.5 & 24.7 & 23.7 & 13.2 & 27.0 & 24.9 & 15.9 & 31.9 & 29.1 & 20.7 \\
			DI-FGSM & 15.0 & 16.2 & 7.1 & 14.7 & 17.7 & 8.4 & 21.6 & 21.1 & 12.9 & 19.3 & 20.2 & 12.7 \\
			TI-FGSM & 30.8 & 30.6 & 22.7 & 30.2 & 31.3 & 23.2 & 36.6 & 36.1 & 29.5 & 36.3 & 36.0 & 30.4 \\
			PI-FGSM(Ours) & \textbf{39.3} & \textbf{39.5} & \textbf{28.8} & \textbf{40.3} & \textbf{41.8} & \textbf{31.4} & \textbf{43.0} & \textbf{45.0} & \textbf{34.9} & \textbf{46.4} & \textbf{48.4} & \textbf{42.4} \\
			\midrule
			TPI-FGSM(Ours) & 49.1 & 50.2 & 36.5 & 49.6 & 51.7 & 38.5 & 51.5 & 51.4 & 43.3 & 59.2 & 60.3 & 56.5 \\
			DPI-FGSM(Ours) & 40.9 & 41.6 & 31.2 & 43.4 & 46.0 & 33.8 & 47.9 & 48.5 & 41.4 & 48.8 & 50.4 & 45.8 \\
			DTMI-FGSM & 50.7 & 50.3 & 39.5 & \textbf{54.0} & 54.2 & \textbf{42.9} & 59.5 & 58.4 & 53.1 & 65.4 & 63.8 & \textbf{62.9} \\
			DTPI-FGSM(Ours) & \textbf{51.5} & \textbf{53.1} &\textbf{40.0} & 52.2 & \textbf{54.7} & 41.7 & \textbf{65.3} & \textbf{64.5} & \textbf{56.3} & \textbf{70.0} & \textbf{67.7} & 62.3 \\
			\bottomrule
	\end{tabular}}%
	\label{table2}%
\end{table}%

\begin{table}[t]
	\centering
	\caption{The success rate(\%) of non-targeted attacks. We use an ensemble of Inc-v3, Inc-v4, Res-152, and IncRes-v2 to generate our adversarial examples by FGSM, I-FGSM, MI-FGSM, DI-FGSM, TI-FGSM, PI-FGSM, and their combined versions respectively.}
	\resizebox{0.7\linewidth}{!}{
		\begin{tabular}{c|c|c|c|c|c|c}
			\toprule
			\toprule
			& \multicolumn{1}{l|}{Inc-v3$_{ens3}$} & \multicolumn{1}{l|}{Inc-v3$_{ens4}$} & \multicolumn{1}{l|}{IncRes$_{ens}$} & \multicolumn{1}{l|}{ResNeXt$_{DA}$} & \multicolumn{1}{l|}{Res152$_B$} & \multicolumn{1}{l}{Res152$_D$} \\
			\midrule
			\midrule
			FGSM  & 27.1 & 24.0 & 13.5 & 3.1  & 1.4 & 2.2 \\
			I-FGSM & 26.2 & 25.2 & 16.0  & 0.7 & 0.8  & 0.4 \\
			MI-FGSM & 51.9 & 49.3 & 32.9 & 2.8 & 1.8 & 2.2 \\
			DI-FGSM & 40.5 & 38.5 & 25.6 & 1.2 & 1.5 & 1.2\\
			TI-FGSM & 39.3 & 38.9 & 31.5 & 6.1 & 3.7 & 3.0 \\
			PI-FGSM(Ours) & \textbf{61.0} & \textbf{62.8} & \textbf{51.3} & \textbf{8.7} & \textbf{8.5} & \textbf{6.0} \\
			\midrule
			TPI-FGSM(Ours) & 79.6 & 81.4 & 74.0 & 11.5 & 10.5 & 9.4 \\
			DPI-FGSM(Ours) & 66.7 & 68.5 & 58.7 & 9.0 & 7.4 & 6.4 \\
			DTMI-FGSM & 81.2 & 81.1 & 76.6 & 6.1 & 5.5 & 4.8 \\
			DTPI-FGSM(Ours) & \textbf{89.3} & \textbf{89.2} & \textbf{83.4} & \textbf{11.7} & \textbf{10.6} & \textbf{10.4} \\
			\bottomrule
	\end{tabular}}%
	\label{table3}%
\end{table}%

Furthermore, when using Inc-v3 to attack against FD, we are surprised to find that sometimes I-FGSM even perturbs misclassified images into correctly classified ones (See Fig. \ref{fig3}). This may be due to the significant difference in decision boundaries between the NT and FD. Since FD are very robust, and the transferability largely depends on the substitute models. To make our proposed PI-FGSM more convincing, we generate adversarial examples for ResNeXt$_{DA}$, Res152$_B$, Res152$_D$ and an ensemble of them respectively. In this case, we set $\beta=2.5, \gamma = 1$.
\begin{table}[t]
	\centering
	\caption{The average success rate(\%) of non-targeted attacks. The top row models are substitute models (``*'' indicates white-box attack). We use ResNeXt$_{DA}$, Res152$_B$, Res152$_D$ and an ensemble of them to generate adversarial examples by FGSM, I-FGSM, DI-FGSM, TI-FGSM, MI-FGSM, PI-FGSM, and their combined versions respectively.}
	\resizebox{\linewidth}{!}{
		\begin{tabular}{c|c|c|c|c|c|c|c|c|c|c|c|c}
			\toprule
			\toprule
			& \multicolumn{3}{c|}{ResNeXt$_{DA}$} & \multicolumn{3}{c|}{Res152$_B$} & \multicolumn{3}{c|}{Res152$_D$} & \multicolumn{3}{c}{Ensemble} \\
			\midrule
			\midrule
			Attacks & ResNeXt$_{DA}$ & Res152$_B$ & Res152$_D$ & ResNeXt$_{DA}$ & Res152$_B$ & Res152$_D$ & ResNeXt$_{DA}$ & Res152$_B$ & Res152$_D$ & ResNeXt$_{DA}$ & Res152$_B$ & Res152$_D$ \\
			\midrule
			DI-FGSM & 58.3*  & 34.4  & 33.3  & 33.5  & 57.3*  & 32.5  & 34.7  & 34.8  & 56.8*  & 67.0*  & 66.8*  & 65.9*  \\
			TI-FGSM & 66.7*  & 44.5  & 44.6  & 46.9  & 69.8*  & 44.3  & 48.6  & 46.6  & 65.3*  & 63.8*  & 62.1*  & 63.1*  \\
			FGSM  & 67.2*  & 45.7  & 45.0  & 47.5  & 71.3*  & 45.4  & 49.2  & 46.3  & 68.4*  & 67.2*  & 67.1*  & 65.7*  \\
			DTMI-FGSM & 72.1*  & 51.0  & 52.3  & 52.4  & 74.0*  & 49.1  & 54.4  & 51.8  & 69.8*  & 69.6*  & 68.3*  & 65.4*  \\
			DMI-FGSM & 72.5*  & 50.3  & 50.8  & 49.8  & 76.0*  & 48.3  & 51.7  & 48.9  & 72.2*  & 71.7*  & 71.9*  & 71.3*  \\
			MTI-FGSM & 72.6*  & 50.1  & 52.1  & 51.8  & 77.7*  & 49.7  & 53.5  & 51.2  & 73.8*  & 70.6*  & 71.3*  & 68.4*  \\
			MI-FGSM & 79.0*  & 54.0  & 54.4  & 56.6  & 81.5*  & 55.0  & 57.1  & 54.8  & 78.5*  & 75.3*  & 77.1*  & 75.7*  \\
			I-FGSM & 81.6*  & 55.1  & 55.9  & 56.8  & 83.7*  & 55.9  & 57.7  & 55.8  & 82.5*  & 78.9*  & 79.6*  & 79.8*  \\
			PI-FGSM(Ours) & \textbf{86.9*} & \textbf{62.0} & \textbf{60.8} & \textbf{62.0} & \textbf{88.1*} & \textbf{61.9} & \textbf{62.7} & \textbf{62.4} & \textbf{86.6*} & \textbf{84.8*} & \textbf{82.5*} & \textbf{83.2*} \\
			\bottomrule
	\end{tabular}}%
	\label{table4}%
\end{table}%
In Tab. \ref{table4}, we sort these methods in an ascending order. As we can observe, our approach is superior to other methods by a large margin  for both the white-box and black-box settings. However, existing methods' performance is even worse than I-FGSM, which is a basic iterative method. It also reminds us that when attacking several robust defense models, simply combining with different methods may not be effective.
\section{Conclusions}
We propose a novel patch-wise iterative algorithm -- a black-box attack towards mainstream normally trained and defense models, which differs from the existing attack methods manipulating {pixel}-wise noise. With this approach, our adversarial perturbation patches in discriminative regions will be larger, thus generating more transferable adversarial examples against both normally trained and defense models. Compared with state-of-the-art attacks, extensive experiments have demonstrated the extraordinary effectiveness of our attack. Besides, our method can be generally integrated to any gradient-based attack methods. Our approach can serve as a baseline to help generating more transferable adversarial examples and evaluating the robustness of various deep neural networks. 

%Compared with state-of-the-art attacks, we further improve the success rate by 3.7\% for normally trained models and 9.2\% for defense models on average. 

%

%Specifically, we introduce an amplification factor to the gradient in each iteration, and one pixel's overall gradient overflowing the $\epsilon$-constraint is properly assigned to its surrounding patch by a projection kernel. 
%Our method can be generally integrated to any gradient-based attack models.
%Comparing with the current state-of-the-art attack models, we significantly improve the success rate by 9.2\% for defense models and 3.7\% for normally trained models on average. Experimental results also indicate the strong transferability of our generated adversarial examples. 
%Our approach can serve as a baseline to help generating more transferable adversarial examples and evaluating the robustness of various models.

\section{Acknowledgments}
This work is supported by the Fundamental Research Funds for the Central Universities (Grant No. ZYGX2019J073), the National Natural Science Foundation of China (Grant No. 61772116, No. 61872064, No.61632007, No. 61602049), The Open Project of Zhejiang Lab (Grant No.2019KD0AB05).

%\clearpage\mbox{}Page \thepage\ of the manuscript.

\clearpage

\bibliographystyle{splncs04}
\bibliography{egbib}
\clearpage

\appendix
\section{Appendix}
Here we discuss the influence of project factor $\gamma$ and iteration $T$, we consider 12 models to do these experiments:
\begin{itemize}
\item NT: Inc-v3 \cite{ref_article27}, Inc-v4 \cite{ref_article28}, IncRes-v2 \cite{ref_article28}, Res-152 \cite{ref_article29} and Dense-161 \cite{ref_article30}.
\item EAT \cite{ref_article25}: Inc-v3$_{adv}$, Inc-v3$_{ens3}$, Inc-v3$_{ens4}$ and IncRes$_{ens}$.
\item FD \cite{ref_article26}: ResNeXt$_{DA}$, Res152$_B$ and Res152$_D$.
\end{itemize}
Then we report the results for different project factor $\gamma$ settings in Sec. \ref{gamma} and the influence of different iterations $T$ in Sec. \ref{T}.

\subsection{Selection of project factor $\gamma$}
\label{gamma}
In this section, we show the results of difference project factor $\gamma$ settings. For the following tables, the amplification factor $\beta$ and project kernel $W_p$ are fixed to the same settings mentioned in Sec. \ref{exp}. As shown in Tab. \ref{tab:PI},\ref{tab:DPI},\ref{tab:MPI},\ref{tab:TPI},\ref{tab:DMPI},\ref{tab:FD-PI}, setting the project factor $\gamma$ to $\epsilon/T\cdot \beta$ is the best choice in most cases. Besides, it is better not to use large project factor for attack methods with input diversity strategy, and if we use Res152$_B$ as our substitute model to attack against feature denoising models, setting $\gamma$ to 1.0 is better.

% Table generated by Excel2LaTeX from sheet 'gamma'
\begin{table}[htbp]
	\centering
	\caption{The average success rate(\%) of non-targeted attacks in different project factor $\gamma$ settings. Here we use Inc-v3 to generate adversarial examples by PI-FGSM.}
	\resizebox{\linewidth}{!}{
	\begin{tabular}{c|c|c|c|c|c|c|c|c|c|c}
		\toprule
		\toprule
		Inc-v3 & $\gamma$ & Inc-v4 & Res-152 & IncRes-v2 & Dense-161 & Inc-v3$_{adv}$ & Inc-v3$_{ens3}$ & Inc-v3$_{ens4}$ & IncRes$_{ens}$ & Avg \\
		\midrule
		\midrule
		\multirow{16}[1]{*}{PI-FGSM} & 1.0     & 39.2  & 28.7  & 33.8  & 36.5  & 24.2  & 13.7  & 12.8  & 7.1  & 24.5  \\
		& 2.0     & 42.1  & 29.6  & 37.9  & 38.2  & 25.2  & 13.5  & 14.0  & 8.4  & 26.1  \\
		& 3.0     & 42.1  & 28.2  & 35.2  & 38.9  & 25.0  & 16.0  & 14.2  & 8.0  & 26.0  \\
		& 4.0     & 42.8  & 29.9  & 39.7  & 40.8  & 24.9  & 15.6  & 14.8  & 9.0  & 27.2  \\
		& 5.0    & 46.1  & 31.3  & 40.3  & 42.7  & 26.1  & 16.4  & 17.0  & 8.1  & 28.5  \\
		& 6.0     & 48.4  & 31.8  & 40.9  & 44.4  & 27.1  & 18.1  & 16.7  & 9.8  & 29.7  \\
		& 7.0     & 50.9  & 34.7  & 44.0  & 45.5  & 26.5  & 18.8  & 17.9  & 10.0  & 31.0  \\
		& 8.0     & 53.1  & 36.2  & 45.5  & 46.6  & 28.5  & 20.2  & 19.7  & 11.7  & 32.7  \\
		& 9.0     & 52.8  & 35.3  & 45.3  & 49.8  & 28.7  & 21.6  & 21.7  & 13.3  & 33.6  \\
		& 10.0    & 53.2  & 36.9  & 48.8  & 52.5  & 28.5  & 22.1  & 23.3  & 14.8  & 35.0  \\
		& 11.0    & 55.8  & 39.9  & 49.6  & 53.4  & 30.0  & 26.7  & 27.2  & 17.0  & 37.5  \\
		& 12.0    & 55.7  & 40.9  & 50.8  & 55.3  & 29.8  & 28.1  & 29.7  & 18.7  & 38.6  \\
		& 13.0    & 57.4  & 41.7  & 50.3  & 55.6  & 32.1  & 31.2  & 31.5  & 22.7  & 40.3  \\
		& 14.0    & 57.8  & 44.7  & 51.4  & 58.3  & 35.3  & 34.1  & 35.0  & 25.8  & 42.8  \\
		& 15.0    & 58.5  & 44.3  & 52.3  & 59.2  & 38.6  & 37.1  & 36.8  & 26.9  & 44.2  \\
		& 16.0    & \textbf{59.8} & \textbf{46.9} & \textbf{52.7} & \textbf{60.2} & \textbf{40.7} & \textbf{40.2} & \textbf{39.7} & \textbf{29.6} & \textbf{46.2} \\
			\bottomrule
	\end{tabular}}%
	\label{tab:PI}%
\end{table}%

% Table generated by Excel2LaTeX from sheet 'gamma'
\begin{table}[htbp]
	\centering
	\caption{The average success rate(\%) of non-targeted attacks in different project factor $\gamma$ settings. Here we use Inc-v3 to generate adversarial examples by DPI-FGSM.}
	\resizebox{\linewidth}{!}{
		\begin{tabular}{c|c|c|c|c|c|c|c|c|c}
			\toprule
			\toprule
			Inc-v3 & $\gamma$ & Inc-v4 & Res-152 & IncRes-v2 & Dense-161 & Inc-v3$_{adv}$ & Inc-v3$_{ens3}$ & Inc-v3$_{ens4}$ & IncRes$_{ens}$ \\
			\midrule
			\midrule
			\multirow{16}[2]{*}{DPI-FGSM} 
			& 1.0     & 72.9  & 48.6  & 64.0  & 49.5  & 24.7  & 15.5  & 14.4  & 8.4  \\
			& 2.0     & 73.0  & 50.4  & 65.2  & 54.7  & 26.2  & 15.4  & 15.6  & 8.6  \\
			& 3.0     & \textbf{73.2} & \textbf{50.9} & \textbf{66.6} & 60.6 & 26.4  & 16.7  & 17.7  & 9.1  \\
			& 4.0     & 70.1  & 49.0  & 62.1  & 60.3  & 25.9  & 17.5  & 15.8  & 9.1  \\
			& 5.0     & 66.0  & 48.7  & 59.4  & 61.1  & 27.8  & 17.3  & 19.1  & 10.6  \\
			& 6.0     & 62.7  & 47.9  & 54.7  & 61.6  & 27.7  & 18.0  & 19.9  & 9.5  \\
			& 7.0     & 61.8  & 46.0  & 52.0  & \textbf{61.8}  & 28.8  & 20.8  & 20.4  & 11.4  \\
			& 8.0     & 58.2  & 45.5  & 47.8  & 59.2  & 30.4  & 21.8  & 23.8  & 12.0  \\
			& 9.0     & 55.2  & 43.7  & 47.8  & 58.7  & 29.9  & 23.2  & 23.4  & 13.0  \\
			& 10.0    & 53.9  & 44.6  & 46.2  & 59.1  & 31.1  & 26.0  & 26.4  & 16.5  \\
			& 11.0    & 53.8  & 44.3  & 44.6  & 57.6  & 31.2  & 26.9  & 29.7  & 17.9  \\
			& 12.0    & 55.2  & 42.5  & 45.3  & 58.7  & 33.3  & 30.4  & 32.7  & 19.0  \\
			& 13.0    & 52.2  & 43.5  & 43.5  & 58.1  & 34.8  & 33.4  & 34.6  & 22.9  \\
			& 14.0    & 55.3  & 41.7  & 40.8  & 59.0  & 37.0  & 35.0  & 35.9  & 26.4  \\
			& 15.0    & 52.1  & 40.9  & 41.1  & 56.8  & 40.8  & 39.8  & 40.1  & 28.6  \\
			& 16.0    & 51.9  & 43.5  & 41.5  & 57.9  & \textbf{42.3} & \textbf{40.9} & \textbf{41.6} & \textbf{31.2} \\
			\bottomrule
	\end{tabular}}%
	\label{tab:DPI}%
\end{table}%

% Table generated by Excel2LaTeX from sheet 'gamma'
\begin{table}[htbp]
	\centering
	\caption{The average success rate(\%) of non-targeted attacks in different project factor $\gamma$ settings. Here we use Inc-v3 to generate adversarial examples by MPI-FGSM.}
	\resizebox{0.65\linewidth}{!}{
	\begin{tabular}{c|c|c|c|c|c|c}
		\toprule
		\toprule
		Inc-v3 & $\gamma$ & Inc-v4 & Res-152 & IncRes-v2 & Dense-161 & Avg \\
		\midrule
		\midrule
		\multirow{16}[2]{*}{MPI-FGSM} 
		& 1.0     & 56.2  & 41.2  & 54.5  & 46.1  & 49.5  \\
		& 2.0     & 57.2  & 40.2  & 54.9  & 48.0  & 50.1  \\
		& 3.0     & 57.6  & 41.2  & 57.0  & 46.0  & 50.5  \\
		& 4.0     & 57.5  & 42.1  & 57.7  & 46.8  & 51.0  \\
		& 5.0     & 58.8  & 42.0  & 56.2  & 47.2  & 51.1  \\
		& 6.0     & 58.7  & 41.9  & 57.4  & 47.5  & 51.4  \\
		& 7.0     & 57.7  & 43.5  & 57.2  & 48.4  & 51.7  \\
		& 8.0     & 59.1  & 42.4  & 56.3  & 50.8  & 52.2  \\
		& 9.0     & 58.7  & 42.4  & 56.3  & 48.6  & 51.5  \\
		& 10.0    & 60.1  & 43.3  & 57.9  & 49.1  & 52.6  \\
		& 11.0    & 60.1  & 43.4  & 59.5  & 51.9  & 53.7  \\
		& 12.0    & 62.6  & 44.4  & 59.5  & 52.6  & 54.8  \\
		& 13.0    & 62.6  & 45.3  & 59.7  & 55.7  & 55.8  \\
		& 14.0    & 64.0  & 47.7  & \textbf{60.9}  & 57.7  & 57.6  \\
		& 15.0    & \textbf{64.1} & 48.1  & 60.4 & 59.0  & 57.9  \\
		& 16.0    & \textbf{64.1} & \textbf{50.2} & 59.4  & \textbf{60.4} & \textbf{58.5} \\
		\bottomrule
	\end{tabular}}%
	\label{tab:MPI}%
\end{table}%

% Table generated by Excel2LaTeX from sheet 'gamma'
\begin{table}[htbp]
	\centering
	\caption{The average success rate(\%) of non-targeted attacks in different project factor $\gamma$ settings. Here we use Inc-v3 to generate adversarial examples by TPI-FGSM.}
	\resizebox{0.65\linewidth}{!}{
	\begin{tabular}{c|c|c|c|c|c|c}
		\toprule
		\toprule
		Inc-v3 & $\gamma$ & Inc-v3$_{adv}$ & Inc-v3$_{ens3}$ & Inc-v3$_{ens4}$ & IncRes$_{ens}$ & Avg \\
		\midrule
		\midrule
		\multirow{16}[1]{*}{TPI-FGSM} 
		& 1.0     & 32.8  & 29.7  & 31.0  & 20.9  & 28.6  \\
		& 2.0     & 32.0  & 31.2  & 31.6  & 22.2  & 29.3  \\
		& 3.0     & 34.0  & 31.4  & 32.2  & 23.0  & 30.2  \\
		& 4.0     & 34.0  & 32.7  & 34.0  & 22.6  & 30.8  \\
		& 5.0     & 34.9  & 34.3  & 34.7  & 23.4  & 31.8  \\
		& 6.0     & 35.3  & 34.7  & 33.6  & 23.1  & 31.7  \\
		& 7.0     & 35.8  & 35.8  & 36.1  & 24.4  & 33.0  \\
		& 8.0     & 37.9  & 36.4  & 37.5  & 25.9  & 34.4  \\
		& 9.0     & 36.4  & 37.0  & 38.3  & 26.5  & 34.6  \\
		& 10.0    & 38.6  & 38.0  & 38.2  & 26.6  & 35.4  \\
		& 11.0    & 38.0  & 40.9  & 40.3  & 27.1  & 36.6  \\
		& 12.0    & 40.4  & 41.6  & 40.2  & 28.9  & 37.8  \\
		& 13.0    & 41.8  & 42.8  & 43.3  & 29.7  & 39.4  \\
		& 14.0    & 43.8  & 44.0  & 44.7  & 32.9  & 41.4  \\
		& 15.0    & 46.6  & 46.6  & 47.0  & 34.8  & 43.8  \\
		& 16.0    & \textbf{50.2} & \textbf{48.8} & \textbf{50.8} & \textbf{35.2} & \textbf{46.3} \\
		\bottomrule
	\end{tabular}}%
	\label{tab:TPI}%
\end{table}%

% Table generated by Excel2LaTeX from sheet 'gamma'
\begin{table}[htbp]
	\centering
	\caption{The average success rate(\%) of non-targeted attacks in different project factor $\gamma$ settings. Here we use Inc-v3 to generate adversarial examples by DMPI-FGSM.}
	\begin{tabular}{c|c|c|c|c|c|r}
		\toprule
		\toprule
		Inc-v3 & $\gamma$ & Inc-v4 & Res-152 & IncRes-v2 & Dense-161 & \multicolumn{1}{c}{Avg} \\
		\midrule
		\midrule
		\multirow{16}[2]{*}{DMPI-FGSM} 
		& 1.0     & 81.2  & 61.8  & 76.0  & 61.1  & 70.0  \\
		& 2.0     & 81.7  & 64.5  & \textbf{77.3}  & 63.6  & 71.8  \\
		& 3.0     & \textbf{81.9} & \textbf{65.1} & \textbf{77.3} & 66.0  & \textbf{72.6}  \\
		& 4.0     & 79.7  & 63.1  & 75.7  & \textbf{67.9} & 71.6  \\
		& 5.0     & 76.3  & 60.6  & 71.7  & 66.0  & 68.7  \\
		& 6.0     & 74.4  & 56.6  & 69.1  & 67.1  & 66.8  \\
		& 7.0     & 69.5  & 55.6  & 64.0  & 63.6  & 63.2  \\
		& 8.0     & 66.1  & 52.9  & 59.1  & 65.4  & 60.9  \\
		& 9.0     & 63.6  & 51.0  & 58.4  & 64.0  & 59.3  \\
		& 10.0    & 62.2  & 51.1  & 56.1  & 63.5  & 58.2  \\
		& 11.0    & 63.6  & 52.0  & 56.7  & 64.0  & 59.1  \\
		& 12.0    & 61.8  & 49.4  & 56.0  & 64.7  & 58.0  \\
		& 13.0    & 62.9  & 50.1  & 55.6  & 64.8  & 58.4  \\
		& 14.0    & 62.2  & 50.5  & 54.1  & 63.4  & 57.6  \\
		& 15.0    & 60.4  & 49.5  & 53.9  & 62.6  & 56.6  \\
		& 16.0    & 61.4  & 49.7  & 54.7  & 63.8  & 57.4  \\
		\bottomrule
	\end{tabular}%
	\label{tab:DMPI}%
\end{table}%

% ResNeXt$_{DA}$ & Res152$_B$ & Res152$_D$
% Table generated by Excel2LaTeX from sheet 'gamma'
\begin{table}[htbp]
	\centering
	\caption{The average success rate(\%) of non-targeted attacks in different project factor $\gamma$ settings. Here we use Res152$_B$ (``*'' indicates white-box attacks) to generate adversarial examples by PI-FGSM.}
	\begin{tabular}{c|c|c|c|c|c}
		\toprule
		\toprule
		Res152$_B$ & $\gamma$ & ResNeXt$_{DA}$ & Res152$_B$ & Res152$_D$ & Avg \\
		\midrule
		\midrule
		\multirow{16}[2]{*}{PI-FGSM} 
		& 1.0     & 62.0 & \textbf{88.6*} & 61.5 & \textbf{70.7} \\
		& 2.0     & \textbf{62.1}  & 88.2*  & 61.6  & 70.6  \\
		& 3.0     & 61.8  & 87.7*  & \textbf{62.0}  & 70.5  \\
		& 4.0     & 61.5  & 87.1*  & 61.6  & 70.0  \\
		& 5.0     & 60.0  & 85.4*  & 58.8  & 68.1  \\
		& 6.0     & 56.8  & 84.0*  & 56.0  & 65.6  \\
		& 7.0     & 55.0  & 82.1*  & 53.4  & 63.5  \\
		& 8.0     & 53.4  & 80.7*  & 50.8  & 61.6  \\
		& 9.0     & 51.9  & 79.6*  & 49.0  & 60.2  \\
		& 10.0    & 50.9  & 78.7*  & 48.8  & 59.5  \\
		& 11.0    & 51.3  & 78.2*  & 48.6  & 59.4  \\
		& 12.0    & 50.8  & 78.0*  & 48.2  & 59.0  \\
		& 13.0    & 50.2  & 77.8*  & 48.0  & 58.7  \\
		& 14.0    & 50.3  & 77.8*  & 47.7  & 58.6  \\
		& 15.0    & 50.2  & 77.7*  & 47.7  & 58.5  \\
		& 16.0    & 50.2  & 77.7*  & 47.7  & 58.5  \\
		\bottomrule
	\end{tabular}%
	\label{tab:FD-PI}%
\end{table}%

\subsection{The number of iteration $T$ }
\label{T}
In this section, we discuss the influence of different iteration $T$. Here step size is set to $\epsilon / T$, amplification factor $\beta$ is set to $T$, and project factor $\gamma$ is set to $\epsilon / T \cdot \beta$. From the results of Tab. \ref{tab:NT},\ref{tab:EAT},\ref{tab:FD}, we observe that when the iteration $T$ exceeds 10, further increasing it will not bring significant improvement. Besides, the computational cost is proportional to $T$. Since our PI-FGSM is based on FGSM, we want to highlight ``fast". Therefore we do not consider a bigger $T$ in Sec. \ref{exp} and just set $T$ to 10.

% Table generated by Excel2LaTeX from sheet 'gamma'
\begin{table}[htbp]
	\centering
	\caption{The average success rate(\%) of non-targeted attacks in different iteration $T$ against NT. Here we use Inc-v3 to generate adversarial examples by PI-FGSM.}
	\begin{tabular}{c|c|c|c|c|c|c}
		\toprule
		\toprule
		Inc-v3 & iteration $T$ & Inc-v4 & Res-152 & IncRes-v2 & Dense-161 & Avg \\
		\midrule
		\midrule
		\multirow{6}[2]{*}{PI-FGSM} 
		& 5     & \textbf{60.9} & 46.3  & \textbf{54.8} & 58.9  & \textbf{55.2} \\
		& 10    & 59.8  & 46.9  & 52.7  & 60.2  & 54.9  \\
		& 15    & 57.3  & 46.3  & 51.2  & 60.1  & 53.7  \\
		& 20    & 57.6  & 46.6  & 50.8  & 60.6  & 53.9  \\
		& 25    & 58.1  & \textbf{47.2} & 49.7  & \textbf{61.3} & 54.1  \\
		& 30    & 58.6  & 46.0  & 50.0  & 60.7  & 53.8  \\
		\bottomrule
	\end{tabular}%
	\label{tab:NT}%
\end{table}%

% Table generated by Excel2LaTeX from sheet 'gamma'
\begin{table}[htbp]
	\centering
	\caption{The average success rate(\%) of non-targeted attacks in different iteration $T$ against EAT. Here we use Inc-v3 to generate adversarial examples by PI-FGSM.}
	\begin{tabular}{c|c|c|c|c|c|c}
		\toprule
		\toprule
		Inc-v3 & iteration $T$ & Inc-v3$_{adv}$ & Inc-v3$_{ens3}$ & Inc-v3$_{ens4}$ & IncRes$_{ens}$ & Avg \\
		\midrule
		\midrule
		\multirow{6}[2]{*}{PI-FGSM} & 5     & 39.4  & 33.6  & 33.8  & 24.7  & 23.0 \\
		& 10    & 40.7  & 40.2  & 39.7  & 29.6  & 27.4 \\
		& 15    & \textbf{40.8}  & 39.7  & 40.0    & 29.0    & 27.2 \\
		& 20    & 40.5  & 40.6  & 41.3  & 30.1  & 28.0 \\
		& 25    & 39.9  & \textbf{41.3}  & \textbf{41.5}  & 29.9  & 28.2 \\
		& 30    & 40.1  & 41.1  & 40.6  & \textbf{31.4}  & \textbf{28.3} \\
		\bottomrule
	\end{tabular}%
		\label{tab:EAT}%
\end{table}%

% Table generated by Excel2LaTeX from sheet 'FD'

% Table generated by Excel2LaTeX from sheet 'FD'
\begin{table}[htbp]
	\centering
	\caption{The average success rate(\%) of non-targeted attacks in different iteration $T$ against FD. Here we use Res152$_B$ (``*'' indicates white-box attacks) to generate adversarial examples by PI-FGSM.}
	\begin{tabular}{c|c|c|c|c|c}
		\toprule
		\toprule
		Res152$_B$ & iteration $T$ & ResNeXt$_{DA}$    &Res152$_B$     & Res152$_D$    & \multicolumn{1}{l}{Avg} \\
		\midrule
		\midrule
		\multirow{6}[2]{*}{PI-FGSM} & 5     & 60.0  & 86.8*  & 60.2  & 69.0  \\
		& 10    & 61.8  & 88.6*  & 61.5  & 70.6  \\
		& 15    & 62.6  & 89.3*  & 61.7  & 71.2  \\
		& 20    & \textbf{62.9} & 89.5*  & 61.9  & \textbf{71.5}  \\
		& 25    & 62.8  & 89.5*  & \textbf{62.1} & \textbf{71.5}  \\
		& 30    & 62.8  & \textbf{89.8*} & 61.5  & 71.4  \\
		\bottomrule
	\end{tabular}%
	\label{tab:FD}%
\end{table}%

\end{document}